\documentclass{article}

\usepackage{arxiv}

\usepackage[utf8]{inputenc} % allow utf-8 input
\usepackage[T1]{fontenc}    % use 8-bit T1 fonts
\usepackage{hyperref}       % hyperlinks
\usepackage{url}            % simple URL typesetting
\usepackage{booktabs}       % professional-quality tables
\usepackage{amsfonts}       % blackboard math symbols
\usepackage{nicefrac}       % compact symbols for 1/2, etc.
\usepackage{microtype}      % microtypography
\usepackage{cleveref}       % smart cross-referencing
\usepackage{lipsum}         % Can be removed after putting your text content
\usepackage{graphicx}
\usepackage[numbers]{natbib}
\usepackage{doi}

\usepackage{multirow}
\usepackage{tabularray}
\usepackage[normalem]{ulem}
\useunder{\uline}{\ul}{}

\title{MatSAM: Efficient Extraction of Microstructures of Materials via Visual Large Model
}

\newif\ifuniqueAffiliation
\ifuniqueAffiliation % Standard variant of author block
\author{ 
        Changtai Li \textsuperscript{†} \\
	Beijing Advanced Innovation Center\\
        for Materials Genome Engineering\\
	University of Science and Technology Beijing\\
	Beijing, China \\
	\texttt{lichangtai17@gmail.com} \\
	\And
        Xu Han \textsuperscript{†}\\
	School of Intelligence Science and Technology\\
	University of Science and Technology Beijing\\
	Beijing, China \\
	\texttt{1286313960@qq.com} \\
        \And
        Xiaojuan Ban* \\
	Beijing Advanced Innovation Center\\
        for Materials Genome Engineering\\
	University of Science and Technology Beijing\\
	Beijing, China \\
	\texttt{banxj@ustb.edu.cn}
}
\else
\usepackage{authblk}

\newbox{\orcid}\sbox{\orcid}{\includegraphics[scale=0.06]{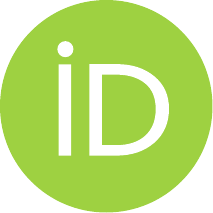}} 
\author[1,2]{Changtai Li\textsuperscript{†}
}
\author[2]{Xu Han\textsuperscript{†}
}
\author[1]{Chao Yao
}
\author[1,2]{Xiaojuan Ban\thanks{Corresponding author: banxj@ustb.edu.cn. \textsuperscript{†}These authors have contributed equally to this work.}
}
\affil[1]{Beijing Advanced Innovation Center for Materials Genome Engineering, University of Science and Technology Beijing, Beijing, China, 100083}
\affil[2]{School of Intelligence Science and Technology, University of Science and Technology Beijing, Beijing, China, 100083}
\fi

\renewcommand{\shorttitle}{\textit{arXiv} Template}

\hypersetup{
pdftitle={A template for the arxiv style},
pdfsubject={q-bio.NC, q-bio.QM},
pdfauthor={David S.~Hippocampus, Elias D.~Striatum},
pdfkeywords={First keyword, Second keyword, More},
}

\begin{document}
\maketitle

\begin{abstract}
Efficient and accurate extraction of microstructures in micrographs of materials is essential in process optimization and the exploration of structure-property relationships. 
Deep learning-based image segmentation techniques that rely on manual annotation are laborious and time-consuming and hardly meet the demand for model transferability and generalization on various source images. Segment Anything Model (SAM), a large visual model with powerful deep feature representation and zero-shot generalization capabilities, has provided new solutions for image segmentation. 
However, directly applying SAM to segmenting microstructures in microscopy images without human annotation cannot achieve the expected results, as the difficulty of adapting its native prompt engineering to the dense and dispersed characteristics of key microstructures in different materials. 
In this paper, we propose MatSAM, a general and efficient microstructure extraction solution based on SAM. A simple yet effective point-based prompt generation strategy is designed, grounded on the distribution and shape of microstructures. 
Specifically, in an unsupervised and training-free way, it adaptively generates prompt points for different microscopy images, fuses the centroid points of the coarsely extracted region of interest (ROI) and native grid points, and integrates corresponding post-processing operations for quantitative characterization of microstructures of materials. 
For common microstructures including grain boundary and multiple phases, MatSAM achieves superior zero-shot segmentation performance to conventional rule-based methods and is even preferable to supervised learning methods evaluated on 16 microscopy datasets whose micrographs are imaged by the optical microscope (OM) and scanning electron microscope (SEM). Especially, on 4 public datasets, MatSAM shows unexpected competitive segmentation performance against their specialist models.
We believe that, without the need for human labeling, MatSAM can significantly reduce the cost of quantitative characterization and statistical analysis of extensive microstructures of materials, and thus accelerate the design of new materials. The source code and data are available at \url{https://github.com/USTB-AI3DVIP/matsam}.
\end{abstract}

% keywords can be removed
\keywords{Material micrograph \and  Microstructure \and Image segmentation \and Visual large model \and Deep learning}

\section{Introduction}

Determining the relationships between microstructures and macro properties has long been a central pursuit in materials science \cite{PropertyPredictionZhiLei2019,MultimodalRelationshipDaRen2023,LinkingAtomicZhenZe2022,ModelAndSim2022}. Micrographs (or microscopy images) provide visual insights into the morphology and spatial distribution of a material's internal structure \cite{durmaz2021deep,uchic3DMicrostructuralCharacterization2006,fengTEMbasedDislocationTomography2020}. Quantitatively characterizing microstructures is therefore crucial for performance prediction, and process optimization \cite{rowenhorstCharacterizationMicrostructureAdditively2020}. Advances in automation – including material experiment robots \cite{szymanskiAutonomousLaboratoryAccelerated2023}, automatic sectioning and grinding machines \cite{echlinSerialSectioningSEM2020} – and imaging technology \cite{hataElectronTomographyImaging2020,fengTEMbasedDislocationTomography2020} have led to an exponential increase in diverse material micrographs. Processing this data often requires extensive human and computational resources. Conventional rule-based image segmentation techniques could suffice for simple, well-imaged micrographs \cite{Canny_1986,Otsu_1979,royAdaptiveThresholdingComparative2014}. To handle diverse and densely packed structures under different imaging conditions, deep learning (DL) has been successfully applied to material micrograph analysis \cite{durmaz2021deep,liDeepMMPEfficient3D2024,HAN2024122110}, and various algorithms \cite{CrackSeg2020} and dedicated software tools \cite{schneiderNIHImageImageJ2012} have emerged.

Not ideally, the effectiveness of DL-based methods hinges on a substantial amount of manually annotated data \cite{ma2020data}. The annotating process is not only time-consuming and laborious but also requires more expertise in materials science compared to natural images. Subjective inconsistencies and erroneous labeling often occur. Another notorious hurdle is those task-specific models' weak generalizability across different materials and microstructures \cite{2022NASAsegmentation}.
Specialist models for micrograph segmentation achieve optimal performances on certain materials and microstructures, restricting their applicability to new images with varying scales, resolutions, and complexities. Introducing operations like image scaling, and cropping also exacerbates the challenges of reusing the models.
To empower the model with generalizability, pre-training followed by fine-tuning, a popularized learning paradigm to transfer the learned knowledge to adapt to the downstream task is extensively utilized \cite{RethinkingPreTrain2019}. By pre-training on a large-scale dataset, models acquire robust initial feature representations. Subsequent fine-tuning with a smaller, task-specific dataset then pushes models to fit the target domain. In the segmentation of micrographs of materials, \cite{2022NASAsegmentation} developed MicroNet, a large microscopy dataset, and implemented a pre-training strategy for material category classification. The pre-trained model, after transfer learning, achieved impressive out-of-distribution segmentation results. Building upon this work, \cite{alrfouTransferLearningMicrostructure2023} employed the highly scalable Vision Transformer (ViT) architecture \cite{ViT} to explore the potential of long-range spatial relationship perception in microscopy images. Nonetheless, both approaches still require extra supervised fine-tuning and lack the ability for practical zero-shot generalizability when facing unseen material micrographs.

\begin{figure}[h]
    \centering
    \includegraphics[width=1\linewidth]{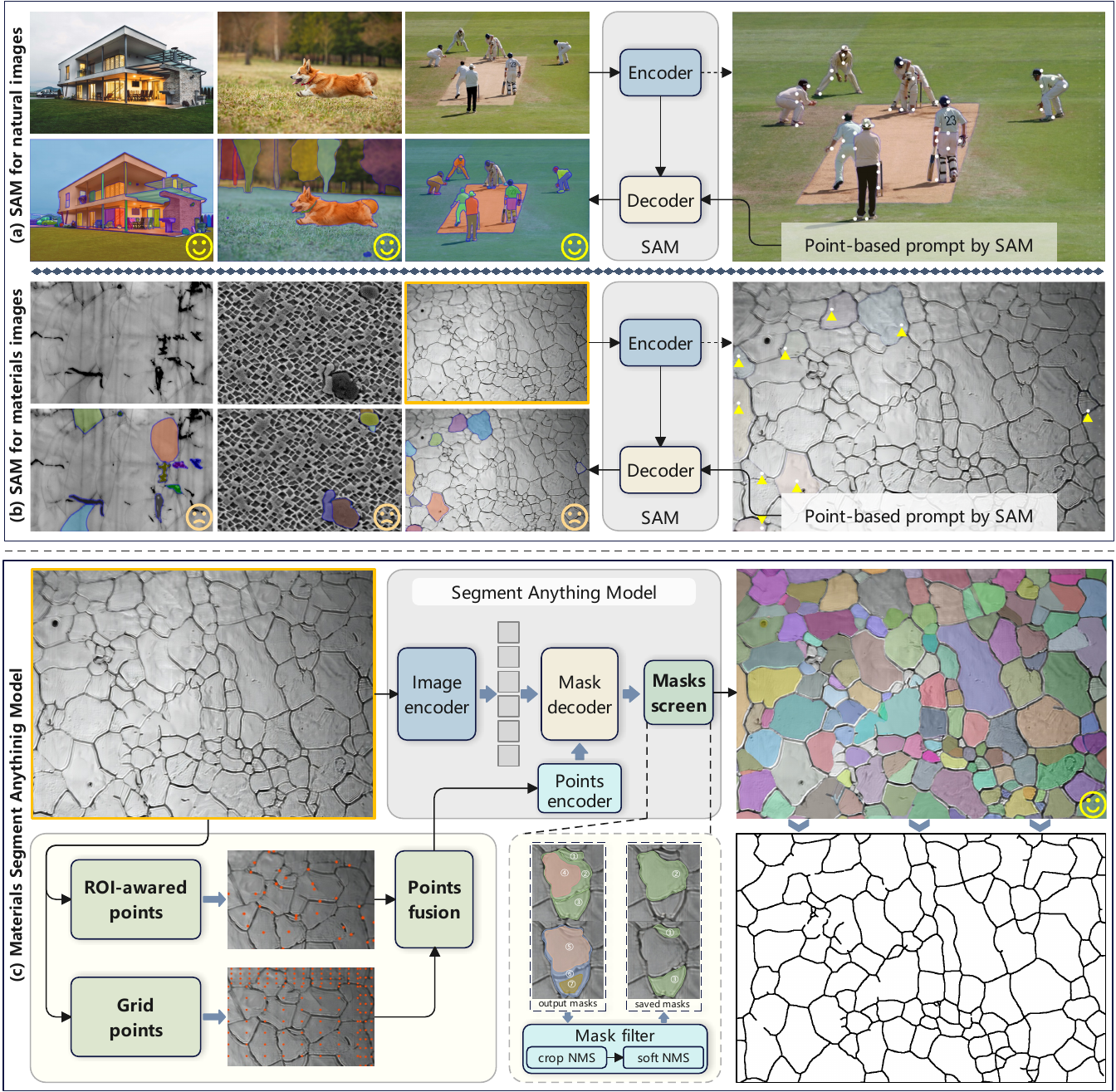}
    \caption{Examples of SAM image segmentation results and the overall architecture of MatSAM. (a) The segmentation results of the native SAM on natural scene images. (b) The segmentation results of the native SAM on three types of material microscopy images. (c) The processing flow and results of MatSAM for material microscopy images.}
    \label{framework}
\end{figure}

Segment Anything Model (SAM) \cite{SAM} emerged as the first generalized visual large model (VLM) for image promptable segmentation, bringing with the zero-shot capability\cite{zhang2023comprehensive}. SAM established both the largest segmentation dataset (at that time) and a corresponding large-scale segmentation model. Trained on billions of image-mask pairs, SAM adopted a simple architecture design and interactive prompt engineering to accurately segment unseen images without additional training (see Fig. \ref{framework} (a)). Prompt engineering tailors the model's behavior through crafted prompts, influencing its output\cite{bommasani2021opportunities}. Consequently, designing scenario-specific prompts becomes crucial for cost-effectively generalizing the VLM to diverse downstream tasks. Existing applications of SAM largely fall into two categories: (1) constructing domain-specific dataset \cite{SAM-cell,MedSAM2024} and retraining/fine-tuning the model \cite{cheng2023sam,autoSAM,HQSAM,all_in_sam}, hindered by huge demand for annotations; or (2) designing task-specific prompt engineering \cite{sam_prompt,chen2023sam}, offering higher flexibility and lower cost. 

In this work, we initially explore the versatility of SAM for material micrographs and assess the native point prompt on diverse microscopy images (see Fig. \ref{framework} (b)). The native approach fails to recognize and segment most microstructures in three types of micrographs (crack, phase, and grain), which can be attributed to SAM's inherent bias toward natural images, i.e., the incompatibility between native point prompts and the characteristics of various microstructures.
Given the complexity and diversity of material microstructures, requiring accurate extraction and generalizability for model reuse, we hereby introduce MatSAM (Materials Segment Anything Model) for segmenting extensive and typical material micrographs. Other than native grid points, MatSAM also employs a structure-aware prompt generation strategy to adaptively deliver prompt points followed by subsequent candidate region screening to yield the final segmentation results. 
Consequently, MatSAM enables unsupervised, training-free, and universal segmentation of microscopy images characterizing typical microstructures of widespread materials, and ultimately achieves accurate and efficient identification and extraction of imaging-agnostic material microstructures.

We evaluate the performance of MatSAM by comparing it with conventional and DL-based (supervised) methods on segmentation results quantitatively and qualitatively. Experiments are conducted on publicly available and in-house datasets of 16 sets of material micrographs, which encompass multiple types of microstructures and are imaged via optical microscopes (OMs) and scanning electron microscopes (SEMs). 
\textbf{Logically}, without manual annotations, MatSAM demonstrates a significant advantage in segmentation accuracy compared to conventional methods. This advantage is particularly pronounced for micrographs with complex and diverse microstructures, as exemplified by the superior performance on eleven datasets of polycrystalline and multiphase materials, including pure iron (PI-1, PI-2), high entropy alloy (HEA), dual-phase steel (DP590-1, DP590-2), etc. 
\textbf{Notably}, without extra training procedure, the accuracies achieved by SAM are on par with, or even surpass typical supervised learning methods \cite{UNET,SEGNET,TransUnet} on two polycrystalline images (PI-1 and SS). 
\textbf{Surprisingly}, On two publicly available datasets with provided test images (nickel-based superalloy (NBS-3\cite{2022NASAsegmentation}) and ultra-high-carbon steel (UHCS\cite{UHCS_and_LCS2023})), MatSAM outperforms the original papers' proposed supervised methods \cite{2022NASAsegmentation}, improving intersection over union (IoU) \cite{IoU} by 3.17\% and 4.8\%, respectively.
The above results confirm the generalizability of MatSAM under zero-shot and training-free conditions. Across different materials, microstructures, and imaging techniques, the accurate segmentation of unseen images with varying visual features and imaging quality facilitates the rapid quantitative characterization of material microstructures.
In light of the competence of microstructure extraction, we believe MatSAM, the first application (to our best knowledge) of the visual large model for analyzing material micrographs can encourage a new research paradigm in the field and thus accelerate the design of new materials.

\section{Results}

A qualitative segmentation comparison between natural images and material microscopy images is conducted, presented in Fig. \ref{framework} (a) and (b). It can be observed that differences lie in the number and distribution of regions of interest (ROIs), few and centered in natural scenes, but dense and dispersed in material micrographs. Given that microstructures have diverse geometric shapes, the overlap between multiple ROIs in microscopy images is much greater.
That could explain the inaccurate and incomplete segmentation of micrographs delivered by native SAM that is even equipped with powerful feature representation ability. 
The observation and conjecture motivate us to investigate the impact of prompt engineering on the efficient generalization of SAM in material micrographs and accordingly come up with MatSAM.

\begin{figure}[h]
    \centering
    \includegraphics[width=1\linewidth]{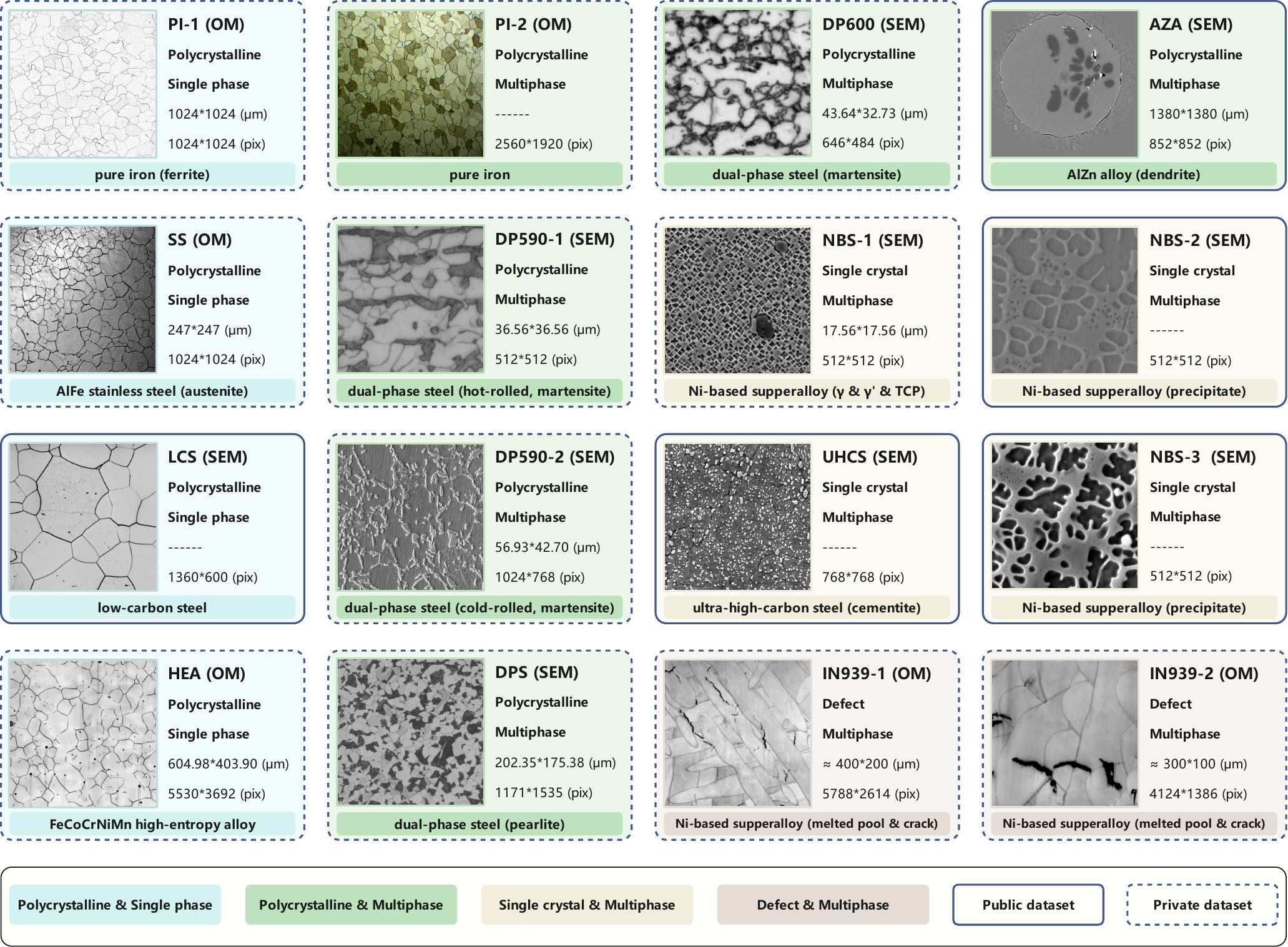}
    \caption{The materials microscopy image dataset used in this paper. The dataset is composed of metal materials, including pure iron, stainless steel, duplex steel, special steel, high-temperature alloy, and high-entropy alloy. The materials are classified into single-phase polycrystalline, multiphase polycrystalline, single-phase polycrystalline, and defects based on the structural features. The imaging methods are divided into optical microscopy and scanning electron microscopy. Best viewed in color.}
    \label{dataset}
\end{figure}

\subsection{MatSAM: Materials Segment Anything Model}

Following a similar procedure as SAM, MatSAM takes the raw image as the input into the encoder, tokenizes the prompts, feeds the learned image feature representation and prompt embedding into the decoder, and outputs the segmentation masks of the target objects. The segmentation results are then used to extract material microstructures, e.g., grains, phases, precipitates, etc. To achieve practical zero-shot and training-free generalizability (segmenting any material microscopy image without human annotations), improving the segmentation prompt is therefore considered.

The native grid-point prompt of SAM provides enough fixed points. Nevertheless, directly using the strategy in material micrographs results in massive unacceptable false segmentations of dense and dispersed target objects. MatSAM adaptively enhances the grid-point prompt with additional generated structure-aware points to profoundly leverage the feature representation capability of SAM with minimal modification. Specifically, conventional image segmentation techniques (Canny \cite{Canny_1986} or OTSU \cite{Otsu_1979}) are first employed to pre-segment the visually distinct target regions. Meanwhile, the regular grid points are dynamically generated according to the distribution of coarsely segmented regions. We argue that providing the model with a meaningful and opportune amount of points is the prerequisite to segmenting the micrograph correctly, without causing redundant computation due to excessive point prompts. By combining the two sets of points, embedded by the prompt encoder, the decoder of SAM upsamples the feature representation of the image to multiple candidate masks. After that, the non-maximum suppression \cite{SOFT-NMS} is operated to screen them for the final segmentation results. A more detailed methodology is described in \ref{methods}.

\subsection{Material micrograph collection and evaluation}

As shown in Fig. \ref{dataset}, we collect 16 material microscopy datasets - 5 public (solid border) and 11 in-house (dashed border) - to evaluate the segmentation performance of MatSAM on different materials and microstructures (four colors to distinguish) under various imaging conditions. The example image, the names of the material and dataset, the type of microstructure, and the real size and resolution of an image are included in each card. Polycrystalline \& Single phase, Polycrystalline \& Multiphase, Single crystal \& Multiphase, and Defect \& Multiphase have blue, green, light yellow, and light brown backgrounds, respectively. All micrographs are imaged via OMs and SEMs. All in-house datasets (except for NBS-1, IN939-1, IN939-2) have manual annotations. To be noted, for public AZA \cite{AZA2020}, NBS-2 \cite{2022NASAsegmentation}, NBS-3 \cite{2022NASAsegmentation} and UHCS \cite{UHCS_and_LCS2023}, we use the attached test data. For LCS \cite{UHCS_and_LCS2023}, because there are no provided labels, part of the raw images are annotated to form the test data. 

As for evaluation, conventional rule-based methods and typical DL models are selected to compare to MatSAM. For polycrystalline images (PI-1, PI-2, SS, HEA, DPS, LCS), grain boundary masks are extracted through image segmentation to calculate the discrepancy between different methods and human labels, and for multiphase images (DP590-1, DP590-2, DPS, DP600, NBS-1, NBS-2, NBS-3, AZA, UHCS), phase masks are directly segmented for comparisons. Importantly, segmentation results by MatSAM are all under zero-shot and training-free conditions.
We conduct extensive quantitative (numerical) and qualitative (visual) evaluations. Conventional methods - Canny \cite{Canny_1986}, OTSU \cite{Otsu_1979}, Watershed \cite{1990MorphologicalWatershed} are implemented for polycrystalline images, and OTSU and adaptive thresholding (Adaptive) \cite{royAdaptiveThresholdingComparative2014} for multiphase images. Supervised DL models (convolutional neural network-based and transformer-based) - UNet \cite{UNET}, SegNet \cite{SEGNET}, and TransUNet \cite{TransUnet} are trained on datasets of relatively high-standard annotations, i.e., PI-1 and SS. Additionally, for public datasets, the segmentation results of MatSAM are directly compared with those of corresponding works.

\subsection{Polycrystalline image segmentation}

\begin{figure}[h]
    \centering
    \includegraphics[width=1\linewidth]{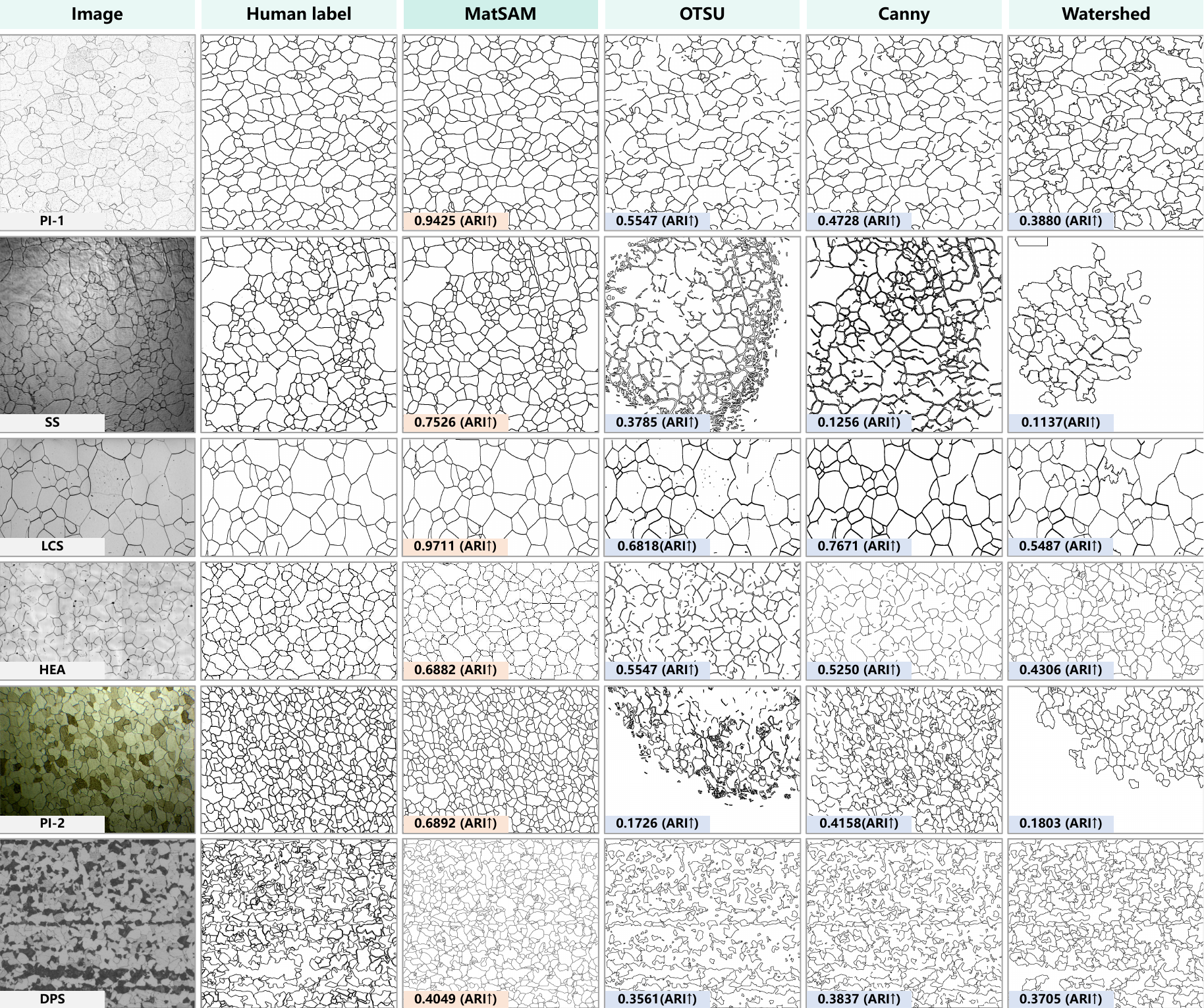}
    \caption{A comparison of MatSAM with conventional rule-based methods on a multi-crystal dataset. The first column shows the original input image, the second column shows the ground truth and the third to sixth columns show the results of MatSAM, OTSU, Canny, and Watershed, respectively. The corresponding segmentation index ARI \cite{ARI} is labeled in each result figure, with higher values indicating better performance.}
    \label{grain}
\end{figure}

Polycrystalline images typically comprise a large number of tightly packed grains, whose boundaries are the target objects of the analysis. As shown in Fig. \ref{grain}, in some micrographs that are imaged via OMs, grayscale values vary little between adjacent grains and largely across regions (SS, HEA, and PI-2, see the first column). MatSAM, after receiving structure-aware point prompts, the reasonable prior knowledge of the target regions to the model, delivers more accurate and refined segmentation results among six datasets (see the third column). On the other hand, threshold methods (OTSU) often suffer from grayscale variations (SS and PI-2, the fourth column). Canny is less affected than threshold methods and can effectively detect most grain boundaries with instinct grayscale differences. However, it is significantly affected by blurred or incomplete grain boundaries (PI-1, SS, and HEA, the fifth column). Following is the quantitative analysis.

\begin{table}[h]
\centering
\caption{Comparison of segmentation results of MatSAM with other supervised methods on publicly available multiphase datasets.}
\label{tab:4}
\begin{tabular}{cccccc}
\hline
\textbf{Dataset}                                                                    & \textbf{Metric} & \textbf{MatSAM}         & \textbf{Canny}        & \textbf{OTSU} & \textbf{Watershed} \\ \hline
\multirow{3}{*}{\begin{tabular}[c]{@{}c@{}}PI-1\\      (w/o postproc)\end{tabular}} & \textbf{ARI↑}   & \textbf{0.8902 {\small (+0.555)} } & 0.3312                & {\ul 0.3348}  & 0.2108             \\
                                                                                    & F1↑             & 0.3762                  & 0.3519                & 0.3541        & 0.2609             \\
                                                                                    & Recall↑         & 0.4985                  & 0.3465                & 0.3261        & 0.5303             \\ \hline
\multirow{3}{*}{\begin{tabular}[c]{@{}c@{}}PI-1\\      (postproc)\end{tabular}}     & \textbf{ARI↑}   & \textbf{0.6336 {\small (+0.108)}} & {\ul 0.5253} & 0.2025        & 0.1979             \\
                                                                                    & F1↑             & 0.7334                  & 0.6460                & 0.2571        & 0.2587             \\
                                                                                    & Recall↑         & 0.8192                  & 0.8668                & 0.7802        & 0.6903             \\ \hline
\multirow{3}{*}{SS}                                                                 & \textbf{ARI↑}   & \textbf{0.5634 {\small (+0.195)}} & {\ul 0.3684} & 0.2049        & 0.1974    \\
                                                                                    & F1↑             & 0.661                   & 0.4984                & 0.3574        & 0.3166             \\
                                                                                    & Recall↑         & 0.5982                  & 0.4650                & 0.3721        & 0.2626             \\ \hline
\multirow{3}{*}{HEA}                                                                & \textbf{ARI↑}   & \textbf{0.5823 {\small (+0.214)} } & 0.3497       & {\ul 0.3682}  & 0.2914             \\
                                                                                    & F1↑             & 0.4726                  & 0.4182                & 0.4069        & 0.3852             \\
                                                                                    & Recall↑         & 0.4154                  & 0.3258                & 0.3349        & 0.3341             \\ \hline
\multirow{3}{*}{DPS}                                                                & \textbf{ARI↑}   & \textbf{0.4516 {\small (+0.017)}} & {\ul 0.4348}          & 0.4059        & 0.3935             \\
                                                                                    & F1↑             & 0.4805                  & 0.5503                & 0.5133        & 0.5114             \\
                                                                                    & Recall↑         & 0.4093                  & 0.4958                & 0.4248        & 0.4363             \\ \hline
\multirow{3}{*}{PI-2}                                                               & \textbf{ARI↑}   & \textbf{0.7525 {\small (+0.389)}} & {\ul 0.3632}          & 0.1392        & 0.1803             \\
                                                                                    & F1↑             & 0.7852                  & 0.5030                & 0.2179        & 0.2524             \\
                                                                                    & Recall↑         & 0.7859                  & 0.7525                & 0.2032        & 0.1904             \\ \hline
\multirow{3}{*}{LCS}                                                                & \textbf{ARI↑}   & \textbf{0.9605 {\small (+0.188)}} & {\ul 0.7730}          & 0.6398        & 0.5406             \\
                                                                                    & F1↑             & 0.9598                  & 0.8311                & 0.6938        & 0.6127             \\
                                                                                    & Recall↑         & 0.9694                  & 0.7692                & 0.7229        & 0.6913             \\ \hline
\end{tabular}
\end{table}

For PI-1, MatSAM achieves an ARI of 0.8908 (without extra post-processing) on 59 test images, while Canny (the second best) only gets 0.3312. When introducing post-processing in favor of conventional methods, MatSAM also obtains superior performance.
The SS dataset is collected under a relatively non-ideal imaging condition, containing more blurry regions, scratches, and impurities, in which an ARI of 0.5634 on 44 test images is obtained, compared to 0.3684 by Canny.  
For HEA micrographs, because every image is stitched by multiple small ones due to the limited field of view, visible horizontal or vertical stitching artifacts are caused by the brightness variations. Hence, segmenting the images of HEA is a challenging task, not to mention that there are a large number of non-closed grain boundaries and protruding cracks. The best segmentation result of the rule-based method (OTSU) is 0.3497 ARI, while our approach gets 0.5823. 
On the PI-2 dataset, 0.7525 (ARI) by MatSAM and 0.3632 by Canny are the top 2 accuracies. Uneven lighting, shown in the fifth row in Fig. \ref{grain} (bottom left and top right areas), brings on the fatal false segmentation by OTSU and Watershed. Even though Canny can handle the situation, its overall segmented grain boundaries are far worse than MatSAM's, numerically and visually.
On the DPS dataset, MatSAM performs slightly better than Canny (0.4516 against 0.4348). The relatively lower value only suggests that the segmentation results are not that consistent with how-quality human labels that have numerous errors (see the second column of the last row in the figure). Nonetheless, It can be seen that MatSAM's segmentation results better fit the grain boundaries in the raw images than the annotations. From this point of view, the objective evaluation metrics sometimes cannot accurately reflect the ability of the approach.
For the public LCS dataset \cite{UHCS_and_LCS2023} with good imaging conditions, MatSAM exhibits the greatest accuracy, achieving an ARI of 0.9605 on 16 test images, which is ahead of other segmentation techniques by a large margin. From the visual segmentation results, MatSAM not only segmented almost all grains in the LCS (see the third row of Fig. \ref{grain}), but also ideally identified the vast majority of incomplete and blurred grain boundaries. In addition, while the image resolution of this dataset varies greatly, MatSAM still outputs satisfactory results, demonstrating its strong robustness.

\begin{table}[h]
\centering
\caption{Comparison of segmentation performance of MatSAM and supervised methods on PI-1 and SS datasets.}
\label{table-supervised}
\begin{tabular}{cccccc}
\hline
\textbf{Dataset}                                                                    & \textbf{Metric} & \textbf{MatSAM} & \textbf{TransUNet} & \textbf{UNet} & \textbf{SegNet} \\ \hline
\multirow{3}{*}{\begin{tabular}[c]{@{}c@{}}PI-1\\      (w/o postproc)\end{tabular}} & \textbf{ARI↑}   & \textbf{0.8902 {\small (+0.259)}}& 0.6317             & 0.5933        & \underline{0.6325}          \\
                                                                                    & F1↑             & 0.3762          & 0.5043             & 0.5648        & 0.5441          \\
                                                                                    & Recall↑         & 0.4985          & 0.5239             & 0.5799        & 0.6242          \\ \hline
\multirow{3}{*}{\begin{tabular}[c]{@{}c@{}}PI-1\\      (postproc)\end{tabular}}     & \textbf{ARI↑}   & 0.6336 & \textbf{0.6957}    & \underline{0.6927}        & 0.6847          \\
                                                                                    & F1↑             & 0.7334          & 0.6453             & 0.7248        & 0.7032          \\
                                                                                    & Recall↑         & 0.8192          & 0.6182             & 0.5719        & 0.5289          \\ \hline
\multirow{3}{*}{SS}                                                                 & \textbf{ARI↑}   & 0.5634 & \textbf{0.6302}    & \underline{0.5943}        & 0.5336          \\
                                                                                    & F1↑             & 0.6610          & 0.7005             & 0.6994        & 0.6443          \\
                                                                                    & Recall↑         & 0.5982          & 0.6313             & 0.6183        & 0.6091          \\ \hline
\end{tabular}
\end{table}
Next, we compare typical DL-supervised methods with MatSAM on the in-house polycrystalline PI-1 and SS datasets. The quantitative results are shown in Table \ref{table-supervised}. Overall, MatSAM attains competitive segmentation results against supervised counterparts.
For PI-1, among supervised methods, SegNet achieves the best ARI of 0.6325 without post-processing, 26\% lower than MatSAM, while TransUNet gets 0.6975 after post-processing (5.91\% higher than our method).
For SS, TransUNet performs best (0.6302 ARI), outperforming MatSAM by about 6.7\%. The classical UNet obtains the second-best value. 
For the datasets containing relatively more image-label pairs, supervised methods can learn iteratively to narrow the distance between segmented masks and abundant human annotations. This allows the model to alleviate the detrimental factors (scratches, impurities, etc.) to some extent. MatSAM, on the other hand, does not undergo the specialist training phase. However, considering the zero-shot and training-free characteristics of MatSAM, such segmentation discrepancies are acceptable, especially under the circumstance of interactive segmentation (human in the loop) \cite{liHALIAHybridActive2023,UHCS_and_LCS2023}.

The robustness and flexibility of our MatSAM should be further highlighted. Dealing with extensive polycrystalline micrographs with varied data preparation protocols, image resolutions, lighting conditions, and many types of defects, MatSAM efficiently and effectively segments almost all target grain boundaries with the combination of structure-aware prior knowledge (supplied by Canny) and capable feature extractor from SAM. On the contrary, supervised methods need task-specific models to receive strong supervision. For the datasets without enough training samples, they are limited, i.e., even after a sufficient number of epochs of training, they may still suffer from performance bottlenecks and overfitting.

\subsection{Multiphase image segmentation}

\begin{figure}[h]
    \centering
    \includegraphics[width=1\linewidth]{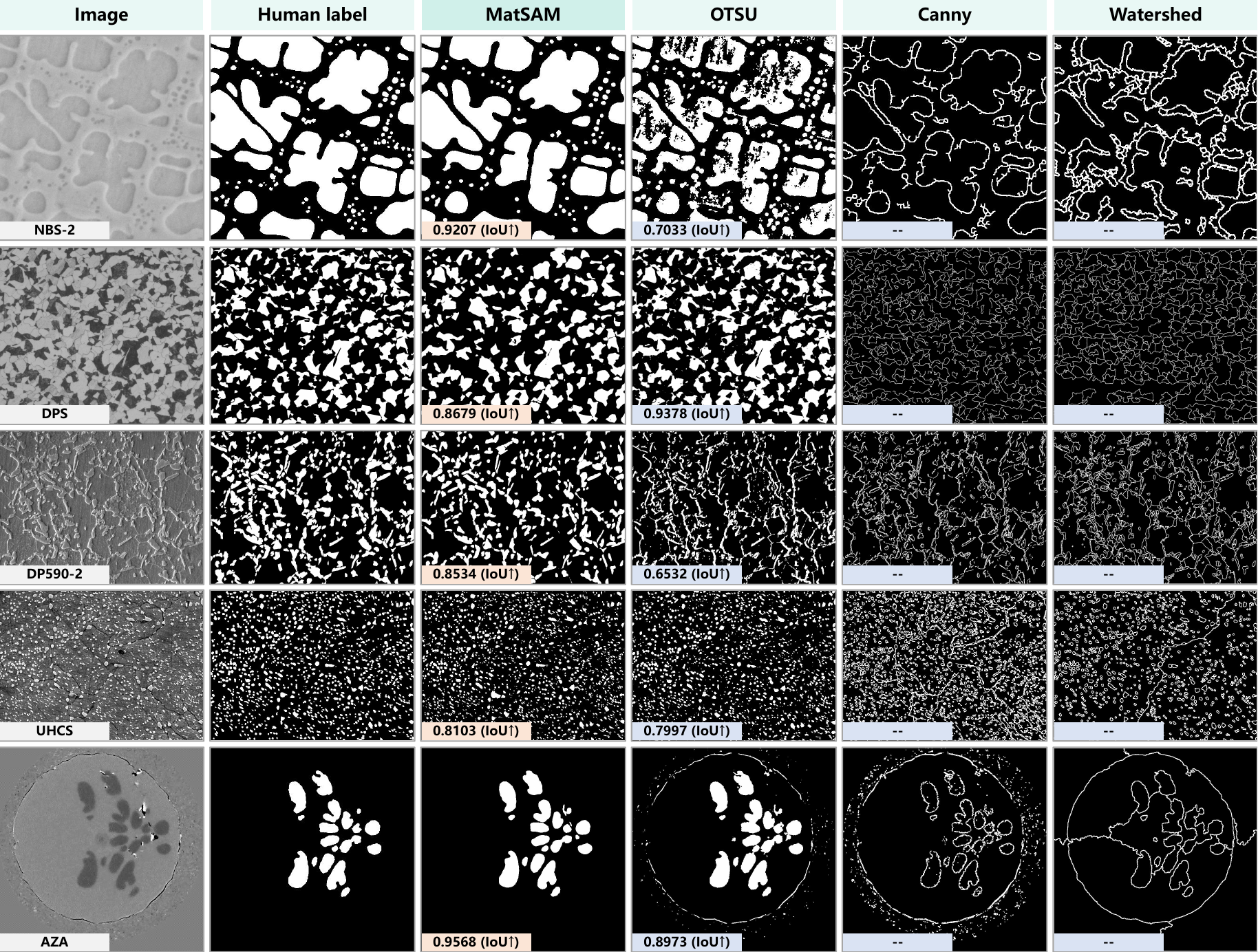}
    \caption{Example of MatSAM versus conventional rule-based methods in a multiphase dataset. The first column shows the original input image, the second column shows the manual annotation, and the third to sixth columns show the output results of MatSAM, OTSU, Canny, and Watershed, respectively. The corresponding segmentation metrics IoU are marked in the results of MatSAM and OTSU (the higher the metric, the better). Since the segmentation results of Canny and Watershed methods are difficult to form effective closed regions, the specific IoU values are not marked.}
    \label{phase}
\end{figure}

\begin{table}[h]
\centering
\caption{Performance comparison of MatSAM with conventional methods on the phase structure dataset.}
\label{table-phase}
\begin{tabular}{ccccc}
\hline
\textbf{Dataset}         & \textbf{Metric} & \textbf{MatSAM} & \textbf{OTSU}   & \textbf{Adaptive} \\ \hline
\multirow{3}{*}{DP590-2} & \textbf{IoU↑}   & \textbf{0.8113 {\small (+0.147)}}& \underline{0.6643}          & 0.2838            \\
                         & F1↑             & 0.8333          & 0.6557          & 0.1659            \\
                         & Recall↑         & 0.8279          & 0.5775          & 0.2158            \\ \hline
\multirow{3}{*}{DP590-1} & \textbf{IoU↑}   & \textbf{0.7811 {\small (+0.075)}}& \underline{0.7061} & 0.5342            \\
                         & F1↑             & 0.8036          & 0.7334          & 0.4971            \\
                         & Recall↑         & 0.8298          & 0.8763          & 0.5198            \\ \hline
\multirow{3}{*}{DPS}     & \textbf{IoU↑}   & \underline{0.8403 {\small (-0.063)}} & \textbf{0.9032} & 0.5178            \\
                         & F1↑             & 0.8802          & 0.9309          & 0.4794            \\
                         & Recall↑         & 0.8493          & 0.9402          & 0.3578            \\ \hline
\multirow{3}{*}{DP600}   & \textbf{IoU↑}   & 0.5760          & \textbf{0.7068} & \underline{0.6190}            \\
                         & F1↑             & 0.5884          & 0.7702          & 0.6671            \\
                         & Recall↑         & 0.5716          & 0.9618          & 0.7488            \\ \hline
\end{tabular}
\end{table}

Micrographs of multiphase microstructures contain one or multiple types of phases (the secondary phase, the precipitate, etc). Unlike grain structures, the phase regions to be segmented do not typically fill the entire image but are irregularly distributed. Also, different phases have conspicuous differences in morphology, structure, and quantity, which can be used to distinguish them after getting the segmented masks.

\begin{figure}[h]
    \centering
    \includegraphics[width=0.8\linewidth]{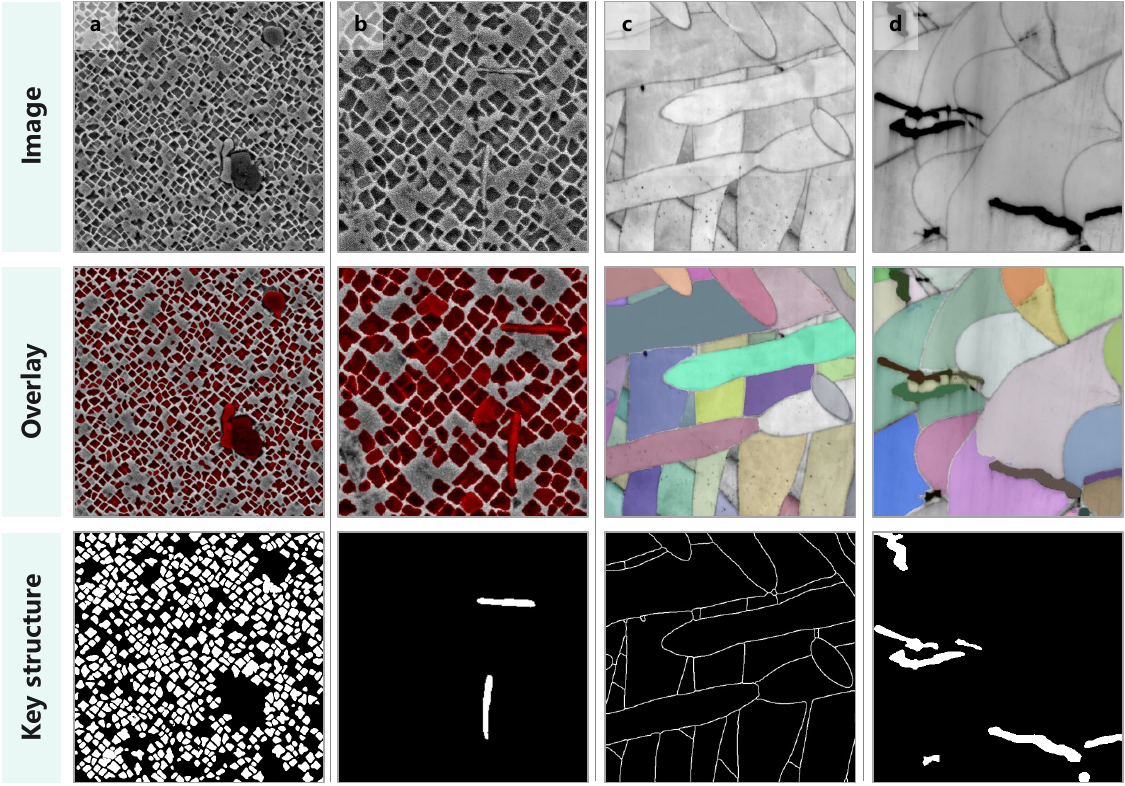}
    \caption{Examples of extracting crucial microstructures in multiphase images. The first row is raw images: (a) NBS-1, (b) NBS-1 (cropped), (c) IN939-1, and (d) IN939-2 (with cracks). The second row is the segmentation masks overlaying the raw images for better visualization. The third row is the binary images highlighting different phases (foreground). Best viewed in color.}
    \label{multiphase_no_label}
\end{figure}

On the two datasets of dual-phase steel datasets (DP590-2, DP590-1), martensite precipitates are the concerned targets, formed during cold and hot rolling, correspondingly. The DP590-2's martensite precipitates appear as light white streaks in the images. The IoU between MatSAM-segmented masks and annotations is 0.8113, observably higher than the OTSU threshold method (0.6643). The DP590-1 dataset consists of 23 hot-rolled martensitic precipitate images with relatively low definition.  The precipitates are of low gray values and are contoured by black boundaries, most of which are more blurred compared to cold-rolled DP590-2 (see Fig. \ref{dataset}), making it more difficult to distinguish. Among the conventional methods, OTSU achieves the better IoU of 0.7061. MatSAM, 7.5\% lower than MatSAM.

For DPS and DP600, MatSAM does not outperform the conventional approaches. From the example image in Fig. \ref{dataset}, DPS's phase regions (foreground) and the background keep a distinct grayscale difference so that given an appropriate threshold value, OTSU can well segment the majority of the darker martensite precipitate phases, with an IoU of 0.9032. MatSAM has a decrease of about 6\%, also outputting acceptable results. When it comes to DP600, MatSAM exhibits a limitation with the image having many uncertain target regions inside which grayscale varies drastically. Another issue comes around the annotation. since there exist unclear tiny regions at the edge of the phase, the inconsistent standard introduced when annotating is not good for the quantitative results of training-free MatSAM, and the issue is discussed in \ref{discussion}.

For multiphase micrographs, multiple phases are to be recognized and extracted. Fig \ref{multiphase_no_label} demonstrates the capability of MatSAM in identifying different phases. For NBS-1, (a) and (b) in the figure, ${\gamma}^{\prime}$ (small light grey areas) and TCP phases (fine needle-like) are the key objects of the analysis. For IN939-1 (c) and IN939-2 (d), melted pools and cracks are of main interest. MatSAM, prompted by structure-aware points, segments nearly all of the mentioned phases. By selecting the candidate regions according to color and morphological features, the final extraction of different microstructures is obtained for further statistical analysis.

\subsection{Evaluating MatSAM on public datasets}

To validate the zero-shot generalizability of MatSAM on more broad scenarios, we perform comparative experiments with conventional approaches and specialist models on 4 material microscopy test datasets released by previous works, including nickel-based superalloy (NBS-2 and NBS-3, corresponding to Super-2 and Super-4 in the original article) \cite{2022NASAsegmentation}, AlZn alloy (AZA) \cite{AZA2020}, and ultra-high-carbon steel (UHCS) \cite{UHCS_and_LCS2023}. Additionally, LCS (low-carbon steel) \cite{UHCS_and_LCS2023} is analyzed qualitatively, for the lack of released labels. Shown in Table \ref{table-public}, compared to specialist models, MatSAM acquires the best results on two datasets, NBS-3 and UHCS, and the second best on AZA, slightly lower than their method.

\begin{table}[h]
\centering
\caption{Comparison of segmentation results of MatSAM with other supervised methods on publicly available multiphase datasets. *Since \cite{2022NASAsegmentation} did not report the numerical results of NBS-2, our method is only compared to OTSU and Adaptive.}
\label{table-public}
\begin{tabular}{cccccc}
\hline
\textbf{Dataset}       & \textbf{Metric} & \textbf{MatSAM   } & \textbf{OTSU}          & \textbf{Adaptive} & \textbf{Supervised method} \\ \hline
\multirow{3}{*}{NBS-2 \cite{2022NASAsegmentation}} & \textbf{IoU↑}   & \textbf{0.9116 {\small (+0.1600)}
}& \underline{0.7516}                 & 0.4458            & *              \\
                       & F1↑             & 0.9544          & 0.8628                 & 0.5192            & /                          \\
                       & Recall↑         & 0.9390          & 0.8692                 & 0.3807            & /                          \\ \hline
\multirow{3}{*}{NBS-3 \cite{2022NASAsegmentation}} & \textbf{IoU↑}   & \textbf{0.8167 {\small (+0.0317)}
}& \underline{0.6678}        & 0.5171            & 0.7850              \\
                       & F1↑             & 0.8760          & 0.7957                 & 0.6239            & /                          \\
                       & Recall↑         & 0.8636          & 0.7873                 & 0.5289            & /                          \\ \hline
\multirow{3}{*}{AZA \cite{AZA2020}}   & \textbf{IoU↑}   & \underline{0.9578 {\small (-0.0112)}} & 0.8853        & 0.5350            & \textbf{0.9690}     \\
                       & F1↑             & 0.9589          & 0.8803                 & 0.3180            & /                          \\
                       & Recall↑         & 0.9595          & 0.9729                 & 0.4777            & /                          \\ \hline
\multirow{3}{*}{UHCS \cite{UHCS_and_LCS2023}}  & \textbf{IoU↑}   & \textbf{0.7584 {\small (+0.048)}
}& 0.5370 \cite{UHCS_OTSU} & 0.4595            & \underline{0.7100}              \\
                       & F1↑             & 0.7445          & /                      & 0.4326            & /                          \\
                       & Recall↑         & 0.7696          & /                      & 0.9276            & 0.8800              \\ \hline
\end{tabular}
\end{table}

\begin{figure}[h]
    \centering
    \includegraphics[width=0.8\linewidth]{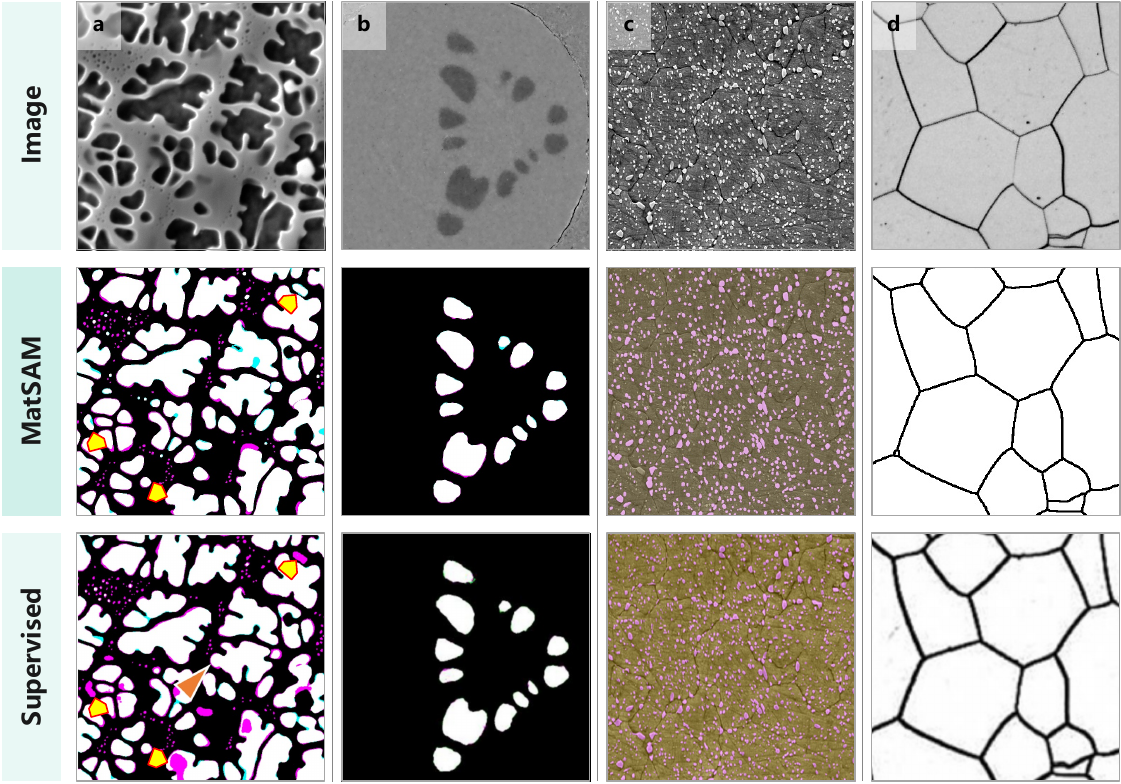}
    \caption{Qualitative segmentation results of MatSAM and other specialist models. The first row is raw open-sourced images: (a) NBS-3 \cite{2022NASAsegmentation}, (b) AZA \cite{AZA2020}, (c) UHCS \cite{UHCS_and_LCS2023}, and (d) LCS \cite{UHCS_and_LCS2023}. The second row is the results of MatSAM, and the third row is the results from data-specific models (the mask images are cropped from the raw figures in their articles, and we are contacting the authors for permission). We inherit from each article the color styles: for results of (a) and (b), pink and blue represent under- and over-segmentation; for (c), pink areas are the masks. Best viewed in color.}
    \label{public_examples}
\end{figure}

NBS-2 and NBS-3 contain multiple precipitates of significantly different sizes. The tertiary precipitates are particularly small and blurry, making segmentation challenging. MatSAM achieves an IoU of 0.9116 on NBS-2, outperforming the conventional methods by a large margin, and 0.8167 on NBS-3, 3.13\% higher than the specialist model. Regular matrix and secondary precipitates are well recognized with few phase contour distortions, and for smaller and blurred tertiary precipitates with low contrast, MatSAM also exhibits better performance, as shown in the first column of Fig. \ref{public_examples} (the yellow arrows mark the regions that MatSAM correctly segments but the original method under-segments). Due to the small size of the dataset (only 4 images), typical supervised methods hardly perform effective training. The authors of the paper conducted pre-training and fine-tuning on specific images to make the model adapt to the target domain. It is worth noting that in the referenced article, the best combination of components of the network was obtained after trying 4 pre-training modes, 40 encoders, and 7 decoder architectures, which is extremely time-consuming and computationally expensive, regardless of the dataset used. Tested on unseen images without fine-tuning, the segmentation performance significantly drops. In contrast, when handling the same diverse unseen images, MatSAM can surpass their results under the zero-shot condition without fine-tuning, underscoring its superior efficient adaptation ability.

AZA is comprised of two XCT images of AlZn dendrite, with the foreground dark precipitate phase in the center of the circular region to segment. For this simple task, dendrites are accurately extracted (see Fig. \ref{public_examples} (b)), and 0.9578 IoU is attained, approaching 0.9690 of the supervised model trained on over 1000 annotated samples. 
Besides, for UHCS, a dataset containing numerous light-colored precipitates (spherical particles) that are uniformly distributed on the top surface of a rough alloy crack. Because of the complex background and large grayscale differences, OTSU performs poorly on this dataset, with an IoU of only 0.5370. MatSAM, on the other hand, gets 0.7584, 4.8\% higher than the specialist model, with the vast majority of cementite particles correctly identified and segmented.
Additionally, for polycrystalline images of LCS (the fourth column of the figure), MatSAM demonstrates a comparable segmentation performance over the original method without human intervention.

To sum up, on the four publicly available test datasets, the comparisons with specialist models highlight the versatility and robustness of MatSAM whose segmentation results are totally usable and even superior either quantitatively or qualitatively. The eminent strengths of unsupervised and training-free make it a universal and efficient solution.

\section{Discussion}
\label{discussion}

\begin{figure}[ht]
    \centering
    \includegraphics[width=0.8\linewidth]{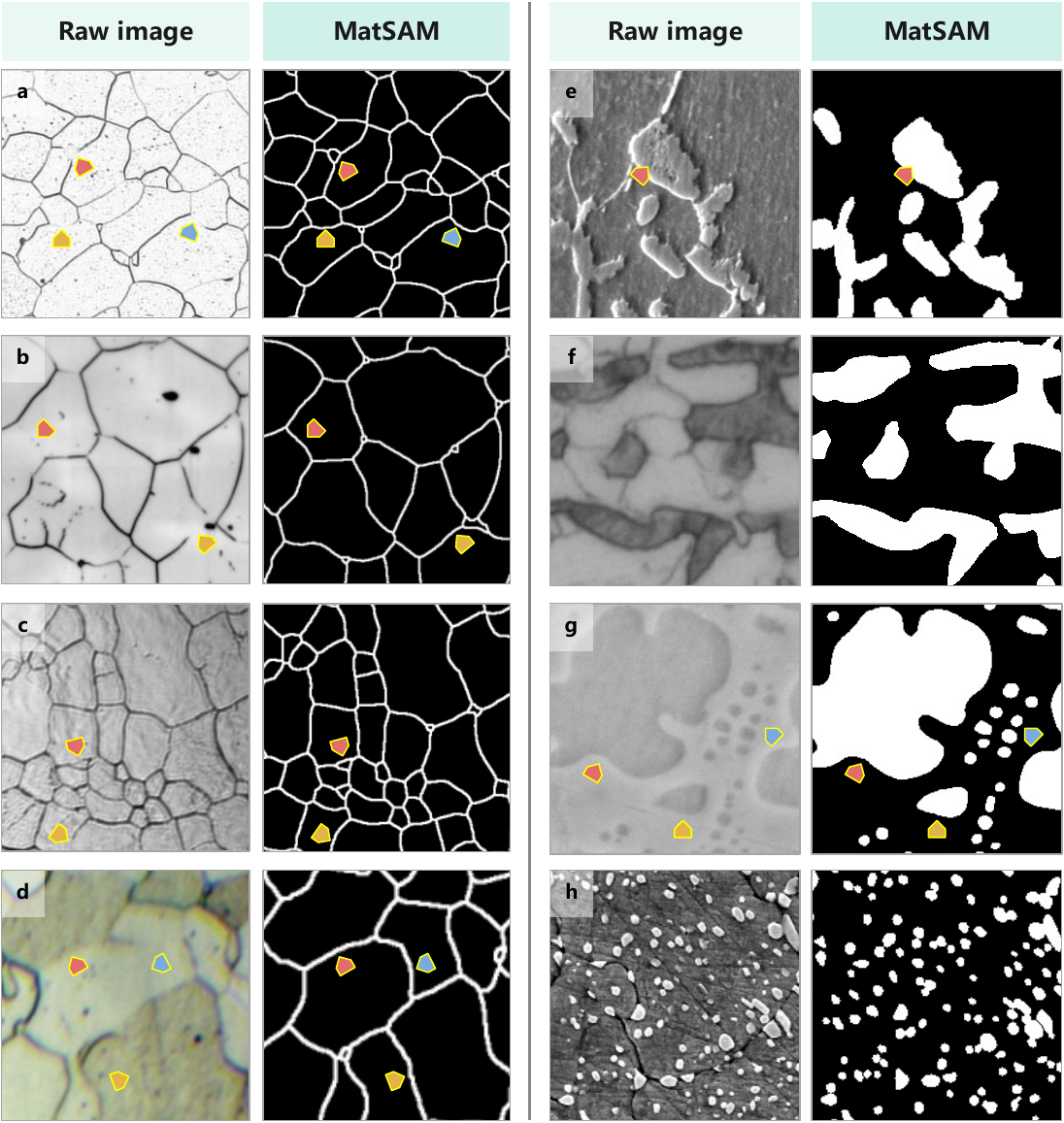}
    \caption{Segmentation results of MatSAM on complicated scenarios. The left side is polycrystalline images, and the right side is multiphase images. (a) PI-1, (b) HEA, (c) SS, (d) PI-2, (e) DP590-2, (f) DP590-1, (g) NBS-2, (h) UHCS. Arrows point to noteworthy areas (a pair of arrows shares the same color). Best viewed in color.}
    \label{complex_structures}
\end{figure}

MatSAM is the first known work to apply the visual large model, SAM \cite{SAM}, to the segmentation of material microscopy images. The core idea of the adaptive structure-aware prompt strategy according to the characteristics of material micrographs, and the further usage of SAM's powerful deep feature representation ability, together shape the effectiveness and efficiency of MatSAM to segment critical and complicated microstructures among extensive materials (see Fig. \ref{complex_structures}).
Abundant experimental results demonstrate its superior capability compared with conventional techniques, typical DL-based methods, and specialist models.
The proposed MatSAM tackles the challenge of high time and labor costs when quantitatively characterizing microstructure. It also further addresses the demand for strong generalizability when reusing the model on unseen images.

Fig. \ref{complex_structures} shows the performance of MatSAM when handling complex structures. For grains in the images, one common issue is that incompleteness or non-closing occurs on the boundary, as marked on the left of the figure. Most untailored approaches underperform in the situation, causing under-segmentations and further detrimentally affecting the statistical accuracy. MatSAM can mitigate this problem, as exemplified in (d); some boundaries are hard to tell at the grain interfaces, nevertheless, MatSAM captures the subtle changes near the boundaries and reasonably recognizes them.
For multiphase images from (e) to (h), phase regions containing small holes (e), target phase boundaries mixed with surrounding areas (f), precipitated sizes differed amongst large, medium, and small (g), and small-sized particles scattered throughout the image (h), MatSAM ideally identify the difficult areas. 

Given its superiority in extracting material microstructures, We believe that MatSAM can be used as a capable tool in the workflow of analyzing material microstructures to accelerate the quantitative characterization based on microscopy images. For normal micrographs, segmentation masks generated by MatSAM can meet the accuracy requirement for subsequent descriptive statistics. In challenging scenarios where imaging conditions are complex, structures are diverse, and so on, MatSAM can replace conventional or DL-based methods as a pre-segmentation model, to give a more reliable initial result. Then, through interactive correction involving human intervention, the false-segmented regions are corrected for model training. For instance, in the case of continuous three-dimensional (3D)-slice images acquired by serial sectioning techniques, 3D objects' characterization results are largely determined by the topological relationship between different sections, thus requiring higher segmentation accuracy as a prerequisite for correct tracking and aggregation. Empirically, it usually takes about 45 minutes to pre-segment and manually correct a microscope image with approximately 300 grains using conventional methods. This process is highly susceptible to the subjective biases of the operator. In contrast, using MatSAM can reduce the segmentation plus correction time to within 5 minutes. The essential reason is that the initial results from MatSAM are already accurate enough to identify and segment the vast majority of grains. The operator simply needs to correct the over- or under-segmentation regions caused by distinct defects or image blurring, less susceptible to visual fatigue and distraction.

\begin{figure}[h]
    \centering
    \includegraphics[width=0.7\linewidth]{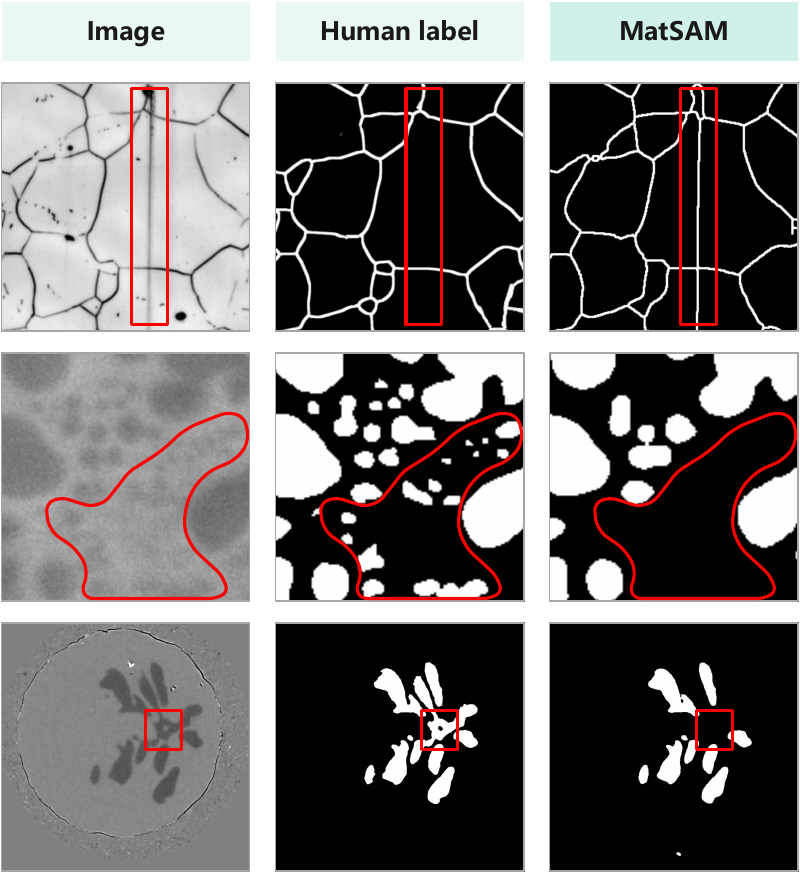}
    \caption{Illustration of under- and over-segmentation by MatSAM on some micrographs. The first column shows the raw image, the second column shows the manual label and the third column shows the MatSAM segmentation results. Best viewed in color.}
    \label{bad_labels}
\end{figure}

MatSAM does have some limitations, as shown in Fig. \ref{bad_labels}.
(1) It still has the sensitivity to imaging defects like scratches and brings with over-segmentation (see the first row of the figure).
(2) When the contrast values are extremely low, for those small objects, MatSAM cannot fully recognize them, the tertiary precipitates in NBS-2 as an example in the figure. It can be attributed to MatSAM having not been trained specifically and cannot recognize textureless regions. In addition, MatSAM misses uncommon structures, as illustrated by the hollow ones in the third row of the figure. That is to say, the prompt points are unable to effectively locate these hard targets, resulting in quite several under-segmentations.
(3) Finally, MatSAM may generate too many prompt points when the input image resolution is too high or the number of ROIs in the field of view is too large, which increases the computational burden of model inference and prediction. 

Correspondingly, we put forward the following possible future improvements for MatSAM: 
(1) Types of defects and their formation mechanisms that are likely to occur in micrographs due to both subjective and objective factors should be summarized and formulated. Explicit physical modeling \cite{Moon2021} and implicit feature representing can be combined to construct additional learnable prompt engineering modules for image defects i.e., to prompt the model to pay attention to regions that may affect the recognition results, further improving the robustness of the model.
(2) Retraining and fine-tuning of MatSAM's parameters based on self-supervised learning (SSL) \cite{SSLVisual2021,rixnerSelfsupervisedOptimizationRandom2022} on a large-scale material microscopy dataset with more extensive materials, microstructures, and imaging modalities should be considered. This enhances the deep feature representation capability of MatSAM for images in the material microscopy domain, enabling more reasonable and accurate feature extraction for more complex and diverse materials microstructures.
(3) Lightweight deployment of the visual large model for material-specific needs under weak computing resource conditions for its broader and easier applications should be conducted. Under the premise of ensuring accurate identification of material microstructure, techniques of pruning, distillation, and model compression of large-scale deep neural networks can retain highly relevant parts for the scenario. accelerating model inference and optimizing the computational efficiency of MatSAM.

\section{Methods}
\label{methods}

\subsection{Datasets and Evaluation metrics}
\label{eval_metrics}

\begin{table}
\centering
\caption{Information of collected 16 public and in-house datasets. *The images of NBS-1, IN939-1, and IN939-2 are unlabeled, and therefore the numerical metrics and results are not given.}
\label{tabel-datasets}
\begin{tblr}{
  cell{1}{7} = {c=3}{},
  hline{1,18} = {-}{0.08em},
  hline{2} = {-}{0.05em},
}
Name    & Source   & Material                & Imaging & Main type    & Test images & Evaluation metrics~ &                  &                  \\
PI-1    & in-house & pure iron               & OM      & grain        & 59             & \textbf{ARI}        & F1               & recall           \\
SS      & in-house & stainless steel         & OM      & grain        & 44             & \textbf{ARI}        & F1               & recall           \\
LCS     & \textbf{public}   & low-carbon steel        & SEM     & grain        & 16             & \textbf{ARI}        & F1               & recall           \\
HEA     & in-house & high-entropy alloy      & OM      & grain        & 38             & \textbf{ARI}        & F1               & recall           \\
PI-2    & in-house & pure iron               & OM      & grain        & 12             & \textbf{ARI}        & F1               & recall           \\
DP590-1 & in-house & dual-phase steel        & SEM     & phase        & 23             & \textbf{IoU}        & F1               & recall           \\
DP590-2 & in-house & dual-phase steel        & SEM     & phase        & 21             & \textbf{IoU}        & F1               & recall           \\
DPS     & in-house & dual-phase steel        & SEM     & grain \& phase  & 9              & \textbf{ARI}        & F1               & recall           \\
DP600   & in-house & dual-phase steel        & SEM     & phase        & 39             & \textbf{IoU}        & F1               & recall           \\
AZA \cite{AZA2020}    & \textbf{public}   & AlZn alloy              & SEM     & phase        & 2              & \textbf{ARI}        & F1               & recall           \\
NBS-1   & in-house & Ni-based supperalloy    & SEM     & phase        & 1024*          & -    & - & - \\
NBS-2 \cite{2022NASAsegmentation}   & \textbf{public}   & Ni-based supperalloy    & SEM     & phase        & 4              & \textbf{IoU}        & F1               & recall           \\
NBS-3 \cite{UHCS_and_LCS2023}  & \textbf{public}   & Ni-based supperalloy    & SEM     & phase        & 5              & \textbf{IoU}        & F1               & recall           \\
UHCS \cite{UHCS_and_LCS2023}   & \textbf{public}   & ultra-high-carbon steel & SEM     & phase        & 5              & \textbf{IoU}        & F1               & recall           \\
IN939-1 & in-house & Ni-based supperalloy    & OM      & phase        & 311*           & -    & - & - \\
IN939-2 & in-house & Ni-based supperalloy    & OM      & phase        & 150*           & -    & - & - 
\end{tblr}
\end{table}

Datasets information is given in \ref{tabel-datasets}, including the name, source, material, imaging technique, main type of microstructure, number of test images, and evaluation metrics. Four metrics are chosen to demonstrate the performance of segmentation: adjusted rand index (ARI, exclusive for polycrystalline images), IoU (exclusive for multiphase images), F1 score, and recall.

\textbf{ARI} \cite{ARI}. The adjusted Rand index (ARI) is a variation of the Rand index (RI), which measures the proportion of instances that are correctly classified in a given instance set, including both positive and negative samples, given by
\begin{equation}
\label{RI}
    \mathrm{RI}=\frac{\mathrm{TP+TN}}{\mathrm{TP+FP+FN+TN}}
\end{equation}
where $\mathrm{TP}$ is the number of true positives, $\mathrm{TN}$ is the number of true negatives, $\mathrm{FP}$ is the number of false positives, and $\mathrm{FN}$ is the number of false negatives. The RI takes values in the range $[0, 1]$, with higher values indicating better performance. 
The ARI penalizes incorrect classifications more heavily than the RI does, namely,
\begin{equation}
\label{ari}
\mathrm{ARI}=\frac{(\mathrm{RI}-E[\mathrm{RI}])}{(\mathrm{max}(\mathrm{RI})+E[\mathrm{RI}])}
\end{equation}
where $E[\mathrm{RI}]$ represents the expected value of the RI. The ARI assigns different weights to correct and incorrect classifications, with a range of $[-1, 1]$. A higher value indicates that the clustering results are more similar to the true state of affairs.

\textbf{IoU} \cite{IoU}. The intersection over union (IoU) or Jaccard index evaluation metric is a commonly used metric for measuring the performance of object detection and image segmentation models, specifically,
\begin{equation}
    \mathrm{IoU}=\frac{\mathrm{TP}}{\mathrm{TP}+\mathrm{FP}+\mathrm{FN}}
\end{equation}
it measures the overlap between the predicted region of the model and the ground truth and is used to assess the model's localization accuracy for the target.
% \[ARI=\frac{(RI-E[RI])}{(max(RI)+E[RI])}\]

% The RI is calculated as shown in Eq. (2), where TP, true positive, is the number of positive samples that are correctly predicted as positive by the model. FP, false positive, is the number of negative samples that are incorrectly predicted as positive by the model. TN, true negative, is the number of negative samples that are correctly predicted as negative by the model. FN, false negative, is the number of positive samples that are incorrectly predicted as negative by the model.

% \[RI=\frac{(TP+TN)}{(TP+FP+TN+FN)} \]

% (2)Recall

% Recall is a metric used to evaluate the performance of a classification model. It measures the fraction of actual positive samples that are correctly predicted as positive, as shown in Equation (3). Recall primarily concerns the model's ability to capture positives, i.e., how well the model finds the actual positives.

% \[Recall=\frac{(TP)}{(TP+FN)} \]

% (3)F1 score

% The F1 score is a commonly used metric for evaluating the performance of classification models, particularly in the case of imbalanced class distributions. It is defined as the harmonic mean of precision and recall, as shown in Equation (4).

% \[F1=\frac{(2×Precision×Recall)}{(Precision+Recall)} \]

% In machine learning, precision is a measure of a classifier's accuracy in predicting positive instances. It is calculated as the ratio of true positives to all predicted positives.

% \[Precision=\frac{(TP)}{(TP+FP)} \]

% (4)IoU

\subsection{Main architecture of MatSAM}

The framework of MatSAM is built upon the original Segment Anything Model (SAM) \cite{SAM} that is trained on a segmentation dataset of over 10 million images and 10 billion masks. In light of the emergent zero-shot capability and flexibility thanks to prompt engineering, SAM is of great potential to be adapted to and applied in a variety of downstream computer vision tasks. MatSAM reuses the main architecture and its pre-trained parameters of SAM, consisting of an image encoder, a prompt encoder, and a segmentation mask decoder.
The image encoder compresses the input image using a pre-trained large-scale vision transformer (ViT), which can handle images of different resolutions, outputting an image embedding of size $C \times H \times W$. The prompt encoder maps input prompt words of different formats to feature embeddings of 256 dimensions. To generate the final masks, the mask decoder updates the image and prompt embeddings in prompt-to-image and image-to-prompt directions using prompt self-attention (SA) and cross-attention (CS). Then, the image embedding is upsampled through the transposed convolution and the output token is mapped through a multilayer perceptron (MLP) to a linear classifier to obtain the probability of the mask foreground. Introducing a new paradigm of promptable segmentation, SAM takes as input a sparse prompt (point, bounding box, text) or a dense one (mask), and generates the corresponding segmentation result. To completely segment all objects in an image, SAM provides a native segmentation method that runs the segmentation process multiple times based on prompt points and filters out better masks by non-maximum suppression (NMS) from excessive objects.

\subsection{Structure-aware prompt point generation}
SAM provides a native point-based prompt, which generates a 32×32 grid of equally spaced points over the entire image. For natural images, the native points generation strategy can provide enough points to consider all ROIs. However, in material microscopy images, key microstructure objects are often densely distributed. Directly using the native strategy often results in problems such as local mis-segmentation of grain or phase regions and loss of a large number of objects at the edge of the image. 

To effectively leverage the strong deep feature representation capability of SAM, MatSAM enhances the original grid-based point prompt strategy by tailoring it to material micrographs with minimal cost to accurately and efficiently identify targets. The main motivation is to mimic humans providing center points for each target, thus automating the process of generating proper prompt points. In a two-step way, it first locates most target objects (ROIs) to be segmented using conventional rule-based methods. Then, the regular grid points are adaptively generated and adjusted according to the distribution of coarsely extracted ROIs to provide as many appropriate numbers of prompt points as possible, avoiding missing those microstructure regions that are not recognized through the former step.

In the first step, for multiphase images, the threshold method (OTSU) is employed to extract phase regions; and for polycrystalline, the edge detection method (Canny) is used to extract the boundary of grains. Then, the separated connected domain regions are obtained using the OpenCV library \cite{opencv_library}, of which the centroid points serve as part of the SAM prompt points. It should be noted that outlier small regions are removed to eliminate the influence of impurities in the image.
In the second step, a suitable number of prompt points are generated according to the number of objects from pre-segmentation. The potential inaccuracy and incompleteness of the pre-segmentation due to the regions with inconspicuous grayscale differences necessitate the extra grid points to compensate for the omission of the centroid points generated by the threshold method.
In addition, a common issue is that the micrograph exhibits uneven quality caused by unstable light or imaging conditions (See SS and PI-2 images in Fig. \ref{dataset}). To address the problem and cover the ROIs near the edge of the image, MatSAM also generates more dense points in the edge regions, enhancing the segmentation performance of incomplete objects at the edges.

\subsection{Implementation details}

MatSAM inherits SAM and first generates multiple rectangular boxes of different sizes and overlap ratios before encoding the image, which is then cropped in correspondence with the boxes for multi-scale segmentation. Cropped images of different scales and the corresponding prompt points are loaded into the model. 
The prompt encoder then encodes the position information of prompt points and learns an embedding vector. After encoding and decoding the image and prompt embeddings, the target masks are output. To deal with the ambiguousness of multiple masks over a single target, the candidate masks (up to three) are ranked in descending order of scores.
SAM first retains the masks with the highest confidence at each inference. 
Then, NMS is performed on the masks generated through multiple prompt points within the cropped frame, retaining non-overlapping segmentation masks within the frame. 
After all the frames are inferred, we integrate all the masks and then perform extra NMS between the cropped frames. Finally, we obtain a set of final segmentation masks with the highest confidence.

In the comparison of the segmentation performance of MatSAM and supervised methods on PI-1 and SS datasets, UNet\cite{UNET}, SegNet\cite{SEGNET}, and TransUNet\cite{TransUnet} are trained from scratch with a random seed. The dataset was split into training, validation, and test sets in a 6:2:2 ratio. All training data in the dataset was randomly horizontally or vertically flipped to improve the model's robustness to image variations. Adam optimizer and GradScaler gradient scaling are used throughout the training process. The learning rate is set to 5e-4. We use Class-balanced weight (CBW) loss\cite{CBWLoss} as the loss function, which is calculated based on the weight map to mitigate the impact of imbalanced classes (boundary and background).  
The training procedure is stopped if the loss value of the validation set in the training remains stable for ten or more epochs, and the network parameters with the minimum loss value during validation are saved for evaluation.
All the above experiments including the proposed MatSAM are conducted on a single Nvidia RTX3090 24 GB GPU using the PyTorch 1.13.1 library \cite{pytorch}.

\bibliographystyle{unsrtnat}
\bibliography{references}  %%% Uncomment this line and comment out the ``thebibliography'' section below to use the external .bib file (using bibtex) .




% 	\bibitem{keshet2016prediction}
% 	Keshet, Renato, Alina Maor, and George Kour.
% 	\newblock Prediction-Based, Prioritized Market-Share Insight Extraction.
% 	\newblock In {\em Advanced Data Mining and Applications (ADMA), 2016 12th International 
%                       Conference of}, pages 81--94,2016.

% \end{thebibliography}
% \nocite{*} %显示所有文献

\end{document}